\documentclass{article}

  \usepackage[main, final]{neurips_2026}

\usepackage[utf8]{inputenc} 
\usepackage[T1]{fontenc}    
\usepackage{hyperref}       
\usepackage{url}            
\usepackage{booktabs}       
\usepackage{amsfonts}       
\usepackage{nicefrac}       
\usepackage{microtype}      
\usepackage{xcolor}         

\usepackage{enumitem}
\usepackage{colortbl}
\usepackage{multirow}
\usepackage{makecell}
\usepackage{wrapfig}
\usepackage{graphicx}
\usepackage{amsmath}
\usepackage{float}

\setcitestyle{numbers,square,comma}

\usepackage[table,dvipsnames]{xcolor}
\definecolor{rowgray}{gray}{0.92}
\definecolor{mypink}{RGB}{218,0,130}
\definecolor{Lavender}{RGB}{230,230,250}
\definecolor{LightPink}{RGB}{255,182,193}
\definecolor{myyellow}{RGB}{255, 242, 204}

\title{LiFi: LiDAR Generation from Multi-View Images via Geometric and Semantic Collaborative Guidance}

\author{%
  Sizhuo Zhou$^{1,2}$, 
  Xiaosong Jia$^{4\dagger}$, 
  Fanrui Zhang$^{1,2}$, 
  Qifeng Li$^3$, \\
  \textbf{Junjie Li}$^{1,2}$, 
  \textbf{Zirui Wang}$^{1,2}$, 
  \textbf{Yukang Feng}$^2$, 
  \textbf{Yu Hong}$^{2,3}$, \\
  \textbf{Shaofeng Zhang}$^3$, 
  \textbf{Wenlong Liao}$^3$, 
  \textbf{Tao He}, 
  \textbf{Juyong Zhang}$^1$, 
  \textbf{Junchi Yan}$^{2,3\dagger}$ \\ \\
  1. University of Science and Technology of China \\
  2. Shanghai Innovation Institute \\
  3. Shanghai Jiao Tong University \\
  4. Fudan University
}

\begin{document}

\maketitle


\begin{abstract}
  Generating LiDAR point clouds from multi-view camera images is a valuable task with applications in controllable data synthesis, cross-modal simulation and various perception tasks. 
  However, most existing methods are not specifically designed for this task, and the quality of their generated LiDAR remains limited. 
  Moreover, these methods typically focus only on semantic cues in images or generate LiDAR from a monocular camera image.
  In this work, we introduce \textbf{LiFi} (\textbf{Li}DAR \textbf{F}rom \textbf{i}mages), a dedicated framework for generating visually and geometrically aligned LiDAR scenes from multi-view camera images.
  LiFi guides the latent diffusion process through two complementary branches: geometry and semantics. 
  The geometry branch constructs scene features from images through a reverse sampling algorithm and an uncertainty-aware geometric encoder combined with DepthAnything3, whereas the semantic branch provides high-level semantic representations via a cross-view interaction module.
  We further introduce a dual-branch balanced classifier-free guidance (CFG) strategy, which enhances the model’s conditional generation capability while preserving the independent learning and collaborative controllability of the two branches. 
  Extensive experiments demonstrate that LiFi outperforms state-of-the-art methods in generating high-fidelity LiDAR scenes.
  The code will be publicly available at: https://github.com/szzhou88/LiFi.
\end{abstract}

\section{Introduction}
\renewcommand{\thefootnote}{} 
\footnotetext{$^\dagger$ Correspondence Authors.} 
\renewcommand{\thefootnote}{\arabic{footnote}}

The rapid progress of autonomous driving (AD) has drawn widespread attention \cite{jia2024bench2drive, jia2023think, jia2023DriveAdapter}. LiDAR sensors, which provide precise 3D structural information, have become a cornerstone of modern AD systems~\cite{liang2022bevfusion,liu2022bevfusion}.  Nevertheless, acquiring large-scale, high-quality LiDAR data could be expensive and the long-tail data captured by consumer devices might not employ LiDAR due to cost control of the product, limiting the robustness and scalability of data-driven methods. 
To mitigate this, pioneering LiDAR generation approaches~\cite{LiDARGen2022, xiong2023ultralidar, hu2024rangeldm} have been introduced, but they mainly focus on unconditional generation and suffer from limited realism because they fail to exploit the diverse contextual cues present in real-world scenes.
Recently, conditional LiDAR point cloud generation~\cite{wu2024text2lidar, xie2024x, li2025uniscene, liang2025lidarcrafter} has emerged, leveraging inputs such as text, maps, and semantic layouts to achieve notable gains in both the photorealism and controllability of synthesized LiDAR scenes.

In this work, we study the conditional generation of LiDAR scenes from multi-view camera images. Cameras, owing to their low cost and ease of deployment, have become a standard sensor in modern autonomous systems.  Generating LiDAR point clouds from such onboard multi-view images therefore holds broad application potential, where it can complement high-quality 3D perception under cost-sensitive sensor setups, and open up new possibilities for cross-modal simulation~\cite{hess2025splatad}, data augmentation~\cite{fang2021lidar}, and sensor completion~\cite{zhou2023bev,fu2019lidar}. However, this formulation remains relatively under-explored.  A major challenge is the cross-modal gap between images and LiDAR. RGB images provide dense 2D appearance information, while LiDAR captures sparse yet precise 3D geometric measurements, making it difficult to establish accurate correspondences between the two heterogeneous representations. Furthermore, although multi-view cameras provide richer observations, different viewpoints also introduce variations in field of view, occlusion patterns, and semantic redundancy. As a result, existing methods~\cite{ran2024towards} have only preliminarily explored images as conditioning inputs, focusing on semantic cues without explicitly enforcing geometric alignment between modalities, and are often limited to monocular inputs that lack comprehensive spatial awareness.

\begin{figure}[t]
  \centering
  \includegraphics[width=\linewidth]{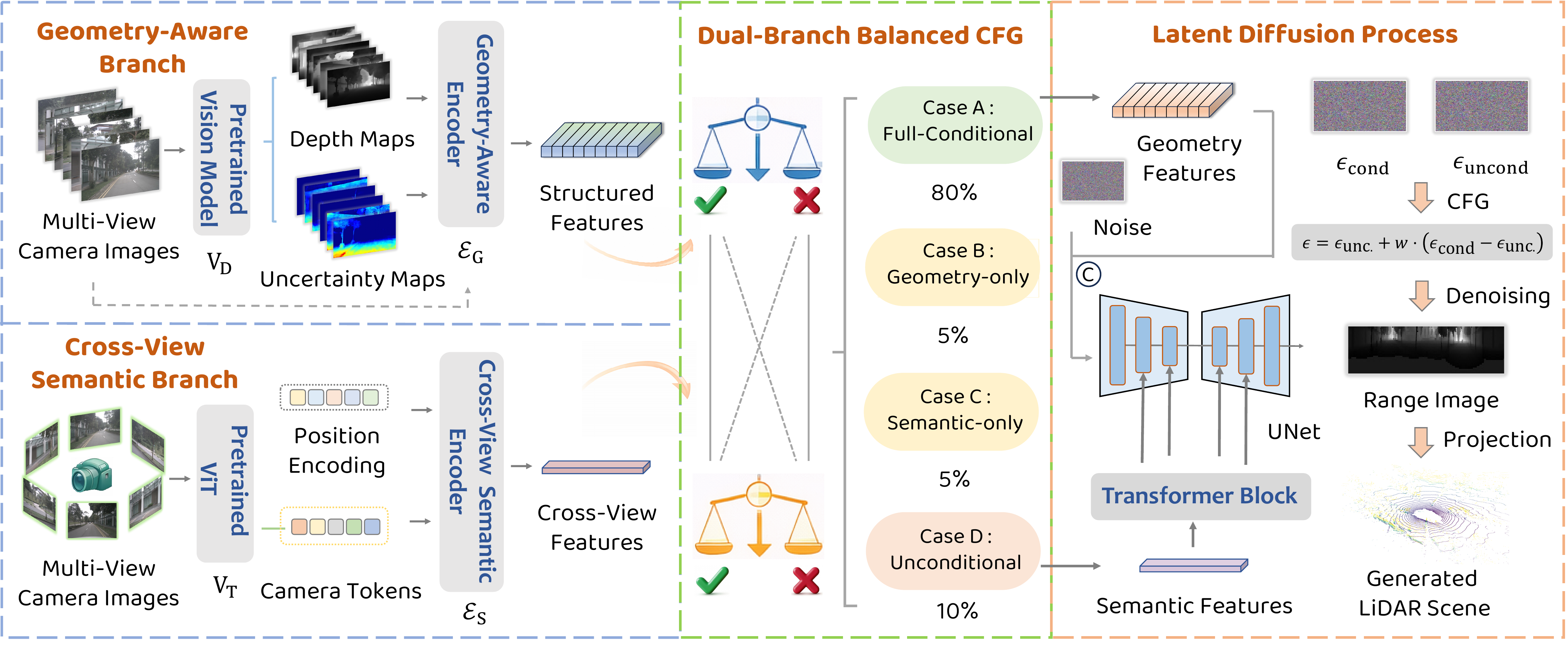}\\
  \caption{\label{overview} 
    \textbf{Overview of the LiFi framework.} The framework takes real multi-view camera images as input and is capable of generating 3D LiDAR scenes with high fidelity that are consistent with the visual information. LiFi mainly comprises the following key designs: (i) geometry–semantic decoupled conditioning branches; (ii) a dual-branch balanced classifier-free guidance module; and (iii) a latent diffusion network integrated with Transformer blocks.
}

\end{figure}

To address these challenges, we propose \textbf{LiFi}, a dedicated framework for generating high-fidelity LiDAR scenes from multi-view camera images.
LiFi guides the latent diffusion process of LiDAR scenes, represented as range images, through two decoupled yet complementary branches: geometry and semantics. In the geometry branch, we first leverage DepthAnything3~\cite{lin2025depth} to estimate per-pixel depth and uncertainty maps from the input images. 
A range-view-centered reverse sampling algorithm then is designed to align these dense predictions, together with the original RGB images, to the LiDAR scanning pattern, producing depth, confidence, and appearance conditions.  These conditions are subsequently encoded into structured geometric features via an uncertainty-modulated encoder that dynamically adjusts feature expressiveness based on local depth confidence. In the semantic branch, we extract high-level visual tokens and fuse them with patch, view, and camera extrinsic encodings. A cross-view attention module then aggregates global semantic context to capture long-range dependencies and cross-view consistency. Furthermore, we introduce a dual-branch balanced classifier-free guidance (CFG) strategy. 
By jointly training with full conditional, single-branch-conditional, and unconditional settings, this strategy not only enhances the model’s conditional guidance capability, but also improves the stability and effectiveness of collaborative learning between the two branches.

We primarily compare LiFi with the state-of-the-art LiDAR generation models on the nuScenes dataset.  
We also compare it with existing frameworks capable of performing camera-to-LiDAR generation. 
To further evaluate the applicability of LiFi, we also compare it with several representative baseline methods on the KITTI-360 dataset \cite{Geiger2013IJRR}.
Experimental results demonstrate the superior performance of our method in generating high-fidelity LiDAR scenes with geometric consistency and semantic alignment. 
Furthermore, we apply LiFi to a downstream object detection task. 
Although this does not directly demonstrate that LiFi-generated scenes can independently support perception performance comparable to that of real LiDAR data, it nevertheless provides evidence of the effectiveness of the proposed generation framework.

In summary, our contributions are as follows:
\begin{itemize}[leftmargin=*]
\item We propose LiFi, which, to the best of our knowledge, is the first dedicated framework capable of generating high-fidelity LiDAR scenes with geometric–visual alignment from multi-view images;

\item We carefully design a range-view-centered reverse sampling algorithm, an uncertainty-aware geometric encoder, and a cross-view semantic encoder to achieve precise cross-modal alignment at the geometric level and effective global semantic guidance;

\item We introduce a dual-branch balanced classifier-free guidance strategy, which effectively enhances the model’s conditional guidance capability and collaborative learning ability;

\item We conduct extensive evaluations of LiFi on the nuScenes dataset for both high-fidelity LiDAR generation and downstream detection tasks, demonstrating the effectiveness of our method.

\end{itemize}


\section{Related Work}

\noindent\textbf{Generative Models.}
Generative models synthesize realistic and diverse new samples by learning the underlying data distribution. 
Early generative frameworks include GANs \cite{goodfellow2014generative} and VAEs \cite{kingma2013auto}. 
GANs can generate realistic images but suffer from issues such as training instability and mode collapse. 
VAEs  provide an explicit probabilistic modeling formulation, but their generated results tend to be overly smooth.
Autoregressive models~\cite{van2016pixel} achieve accurate generation by progressively modeling conditional distributions, yet their sequential sampling process incurs high computational overhead.
Diffusion models (DMs) \cite{sohl2015deep, song2019generative, song2020improved, song2020score, ho2020denoising, nichol2021improved, kingma2021variational, saharia2022photorealistic} have demonstrated outstanding performance in image synthesis.
Subsequent research has introduced latent diffusion models (LDMs) \cite{rombach2022high}, which perform the diffusion process in the latent space, thereby significantly improving generation efficiency.
Meanwhile, they have also facilitated the advancement of conditional generation \cite{nichol2021glide, ramesh2022hierarchical, zhang2023adding}.

\noindent\textbf{LiDAR Scene Generation.}
LiDAR scene generation aims to synthesize 3D point cloud representations with realistic geometric structures and scanning patterns. 
Caccia et al. \cite{caccia2019deep} were the first to adopt the range image representation and perform generation using VAEs and GANs.
Subsequently, LiDARGen \cite{LiDARGen2022} proposed a score-based model for LiDAR generation. 
UltraLiDAR \cite{xiong2023ultralidar} voxelizes LiDAR point clouds and combines this representation with VQ-VAE for generation. 
R2DM \cite{nakashima2024lidar}, built upon DDPMs, improves the fidelity of generated point clouds. 
RangeLDM \cite{hu2024rangeldm} enables fast LiDAR data generation by performing diffusion in a compact latent space.
DriveLiDAR4D \cite{cai2025drivelidar4d} improves LiDAR generation quality by conditioning on three modalities: road sketches, scene descriptions, and object priors. 
UniScene \cite{li2025uniscene} introduces occupancy as an intermediate representation, thereby reducing the difficulty of generating complex LiDAR scenes. 
OpenDWM \cite{ni2025maskgwm} is built upon the DiT architecture for video generation and further extends it to LiDAR scenes.
LiDARCrafter \cite{liang2025lidarcrafter} parses natural language inputs into scene graphs to guide generation. 
LaGen \cite{zhou2025lagen} proposes an autoregressive framework for generating long-horizon LiDAR scenes.

\noindent\textbf{Camera-to-LiDAR Generation.}
Camera-to-LiDAR generation refers to generating 3D LiDAR scenes from camera images. 
Unlike other conditional LiDAR generation models, research on directly synthesizing LiDAR data from camera images remains relatively limited.
LiDM \cite{ran2024towards} proposes a multimodal conditional framework for LiDAR scene generation and also provides a preliminary exploration of the camera-to-LiDAR task by extracting semantic information from camera images. 
However, it does not investigate how to align the image and LiDAR modalities, which differ substantially from each other, and thus its generation quality remains limited.
X-Drive \cite{xie2024x} proposes a framework for the joint generation of LiDAR point clouds and multi-view camera images, taking both LiDAR point clouds and camera images as inputs and generating the two modalities simultaneously. 
It can be adapted to the camera-to-LiDAR task. 
However, its core framework is not specifically designed for the image-only input setting, and therefore does not exhibit clear performance advantages in this scenario. 
Moreover, it incorporates additional text prompts and 3D bounding box information to assist the generation process.
Other frameworks for the joint generation of LiDAR point clouds and camera images include HoloDrive \cite{wu2024holodrive} and Genesis \cite{guo2025genesis}, but they do not provide a camera-to-LiDAR mode.
Our proposed method addresses the above challenges.


\section{Method}

\subsection{Preliminaries}

\noindent\textbf{Range Image Representation.}
We adopt the range image as the basic representation of LiDAR data. Unlike directly processing sparse and unordered 3D point sets, it exploits the sensor’s scanning pattern to encode point clouds into a regular 2D array.
For a point $p=(x,y,z)\in\mathbb{R}^{3}$, we first transform it from the Cartesian coordinate system to a corrected spherical coordinate system \cite{hu2024rangeldm}:
\begin{equation}
\label{corrected_range}
r = \sqrt{x^{2} + y^{2} + (z - h_{j})^{2}}, \quad
\theta = \arctan(y, x), \quad
\phi = \phi_{j},
\end{equation}
where $r$ denotes the depth value, $\theta$ is the azimuth angle, and $\phi$ is the elevation angle; \( h_{j} \) and \( \phi_{j} \) denote the sensor height and elevation angle estimated by Hough voting.
We then map this spherical representation to the range-view space and apply normalization, thereby obtaining the corresponding range image.
This representation transformation is reversible.

Adopting the range image representation not only transforms the data into a 2D representation but also ensures that each pixel corresponds to a specific azimuth and elevation direction of the target LiDAR sensor, thereby preserving the sensor's structured angular sampling topology and inherent sparsity. 
Furthermore, given a calibrated sensor model, the generated 2D representation can be deterministically converted back into a 3D point cloud, improving generation efficiency.

\noindent\textbf{Latent Diffusion Model for LiDAR Generation.}
We adopt a latent diffusion model (LDM) \cite{rombach2022high} for conditional LiDAR data generation. 
The framework consists of a VAE and a UNet: the former compresses the range image into a latent space, while the latter performs diffusion modeling in the latent space.
Specifically, given an input range image $I \in \mathbb{R}^{H_r\times W_r\times 2}$, the encoder first maps it to a latent representation $\hat{z}\in\mathbb{R}^{h_r\times w_r\times c_r}$, and the decoder then reconstructs the range image from this latent variable as $\hat{I}=G(\hat{z})$.
During the diffusion process, the model starts from a clean latent variable $z_{0}$ and progressively injects Gaussian noise into it to obtain a noisy latent variable $z_{t}$.
During training, the UNet takes $z_{t}$, the timestep $t$, and the encoded conditional information $c$ as inputs, and predicts the noise at the current step. 
The training objective is defined as:
\begin{equation}
\label{ldm_loss}
\mathcal{L}_{LDM}=\mathbb{E}_{\epsilon_{t} \in \mathcal{N}(0,1), t \in \mathcal{U}[0, T],c}\left[\left\| \epsilon_{t}-\epsilon_{\theta}\left(z_{t} ; t, c\right)\right\| ^{2}\right],
\end{equation}
where $\epsilon_{\theta}$ denotes the learnable noise prediction network.
During inference, the model starts from random Gaussian noise and iteratively denoises it to generate the latent range-view representation.

\begin{figure}[t]
  \centering
  \includegraphics[width=\linewidth]{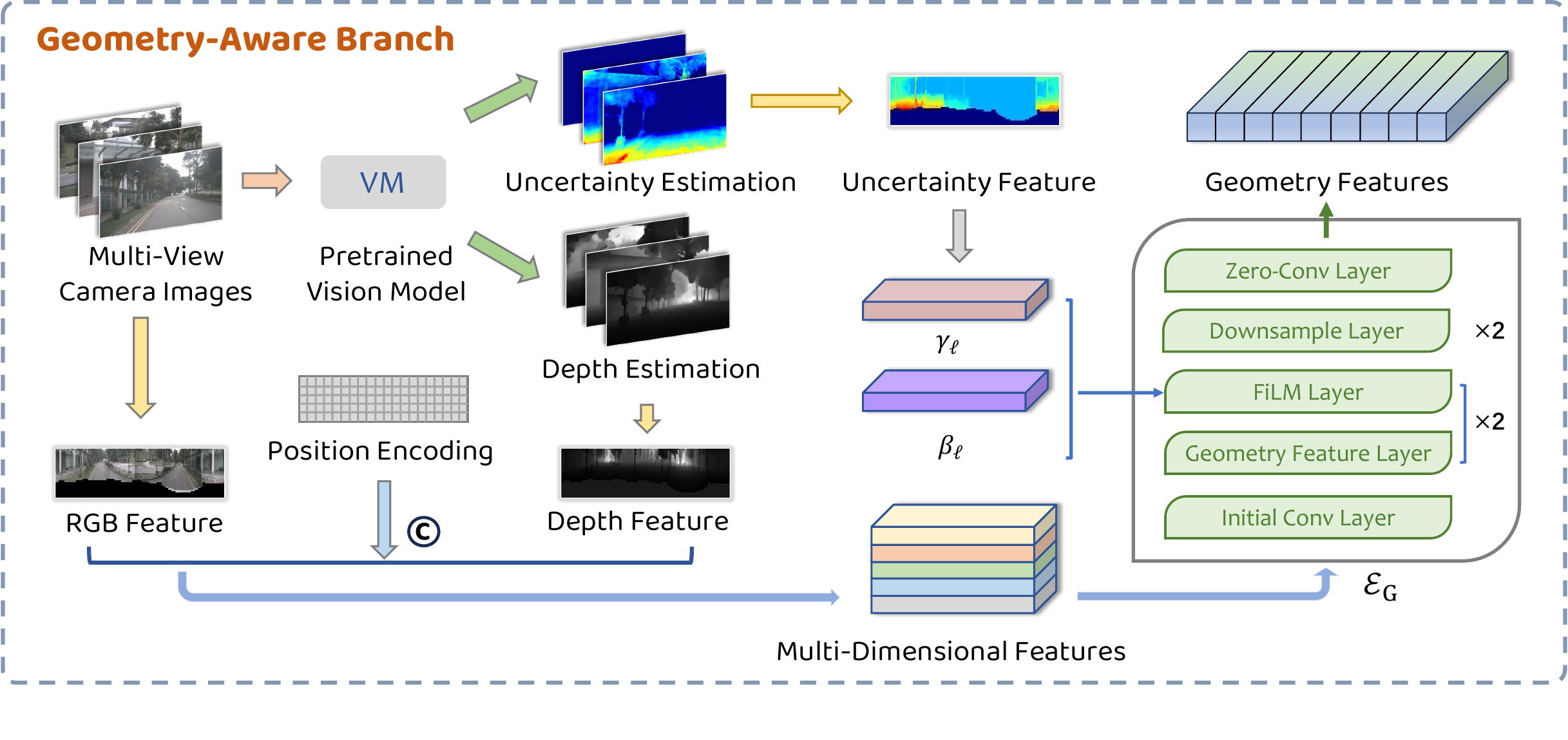}\\
  \caption{\label{geometry_branch} 
    \textbf{Geometry-Aware Branch} extracts structured geometric features from raw images via a range-view-centered reverse sampling algorithm and an uncertainty-modulated geometric encoder, thereby guiding the latent diffusion process.
}
  \end{figure}

\subsection{Overall Framework}
Given the multi-view camera images of a driving scene, $V \in \mathbb{R}^{N_v\times H_v\times W_v\times 3}$, along with the corresponding camera extrinsics $M = [R|t] \in \mathbb{R}^{3 \times 4}$ and camera intrinsics $K \in \mathbb{R}^{3 \times 3}$, our goal is to generate a 3D LiDAR scene that is geometrically and perceptually aligned with the input.
To achieve this goal, we first extract the following three types of visual priors from multi-view camera inputs using a pre-trained vision foundation model: (i) pixel-wise relative depth estimation in the camera image space: $D \in \mathbb{R}^{N_v \times H_v \times W_v}$; (ii) the corresponding uncertainty maps of depth estimation $U \in \mathbb{R}^{N_v \times H_v \times W_v}$; and (iii) camera semantic tokens $S \in \mathbb{R}^{N_v \times l_s \times c_s}$.
Taking DepthAnything3 \cite{lin2025depth} as an example, it can jointly process multiple images captured from different viewpoints of the same scene and natively output corresponding confidence maps.

Based on the above conditional information, we construct an image-to-LiDAR point cloud generation framework guided by dual conditional branches with decoupled geometry and semantics. 
In the first branch, we construct a geometry-conditioned encoder integrated with an uncertainty modulation mechanism to fuse raw RGB images, depth maps, and uncertainty features. 
In the second branch, we build a pose-aware semantic encoder to provide global semantic context and capture cross-view relationships for the model.
The overall pipeline of LiFi is illustrated in Figure \ref{overview}.

\subsection{Reverse Sampling Based Geometry-Aware Conditioning}
In this section, we describe the construction of the geometry-aware conditional branch in detail. 
This branch is designed to extract geometric features of the scene from multi-view camera observations, thereby providing explicit spatial structural constraints for the subsequent diffusion process. 
To effectively encode the geometric details contained in multi-view camera images, we first design a range-view-centered reverse depth sampling algorithm.

\subsubsection{Range-View-Centered Reverse Sampling}
To fully exploit the estimated depth maps $D$ obtained from camera images, a straightforward approach is to project these depth maps pixel by pixel into LiDAR point clouds and then convert them into range images. 
However, this strategy fails to ensure accurate alignment between projected depth maps and LiDAR scenes, while also introducing substantial computational overhead.
\begin{wrapfigure}[15]{r}{0.50\textwidth}
    \vspace{-0.5\baselineskip}
    \centering
    \includegraphics[width=\linewidth]{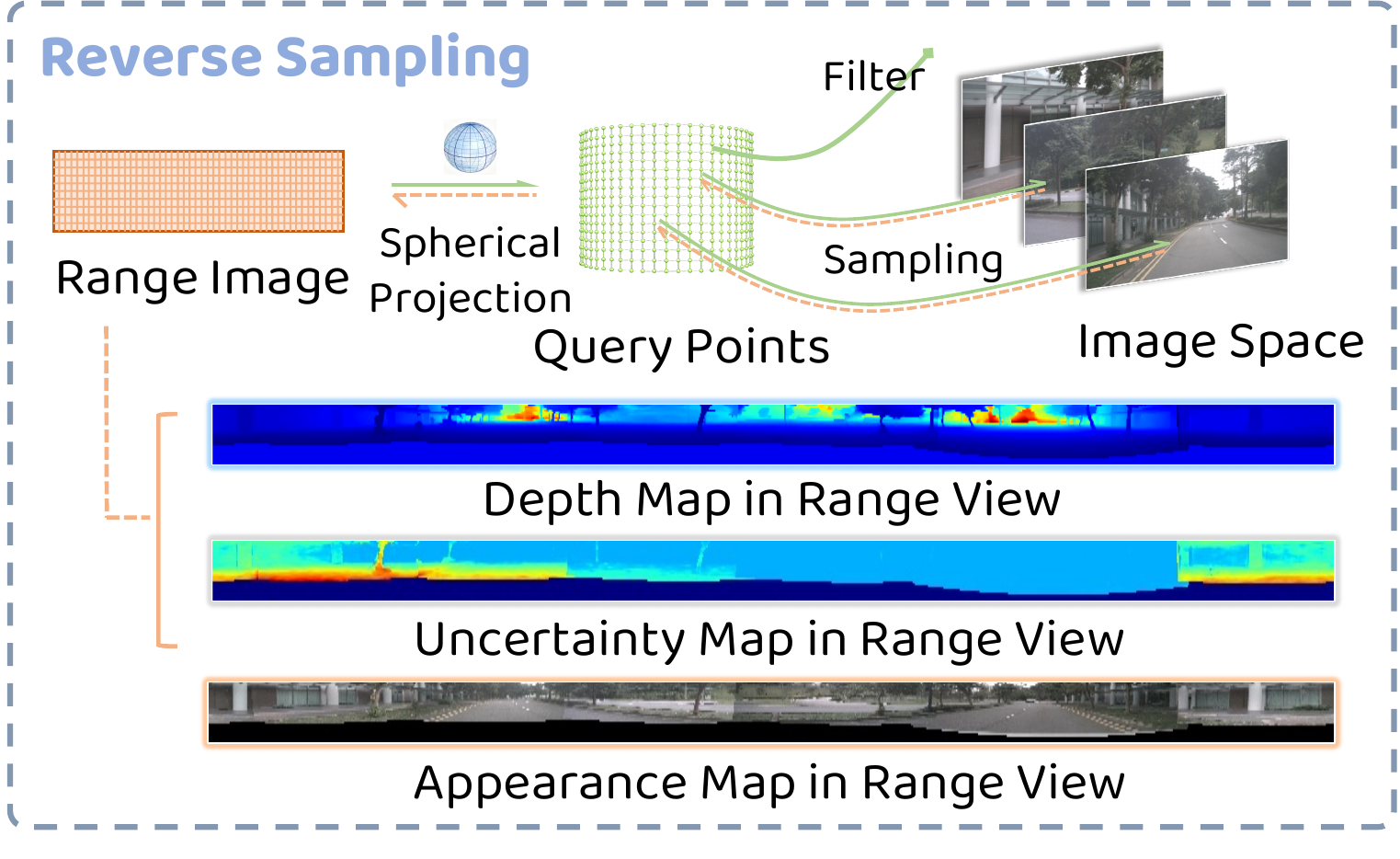}
    \vspace{-6mm}
    \caption{\label{pipeline}
    Schematic illustration of the reverse sampling mechanism.}
\end{wrapfigure}
This issue arises because a single depth map contains millions of pixels (a common resolution is $1600 \times 900$), whereas a typical LiDAR scene is inherently sparse. 
Direct projection is therefore not only inefficient, but also inconsistent with the distribution pattern of LiDAR data. 
To address this, we introduce a range-view-centered reverse depth sampling algorithm.

Specifically, to construct the relative depth map $I_D$ in the range-image representation, we first apply the inverse spherical projection transformation to a standard single-channel range image $I \in \mathbb{R}^{H_r\times W_r}$, yielding a set of canonical 3D query points $Q_D=\{q_k\}$.
We then project these query points, which follow the distribution characteristics of LiDAR scenes, into the camera coordinate systems of different views through a spatial coordinate transformation:
\begin{equation}
\label{coordinate_transformation}
[\tilde{u}_n,\tilde{v}_n,\tilde{w}_n]^T = K_n \cdot (R_n \cdot q_k + t_n).
\end{equation}
Here, $K_n$ denotes the camera intrinsic matrix of the $n$-th image, while $R_n$ and $t_n$ are the rotation and translation parameters of the corresponding camera extrinsics, respectively.
We then project them into the pixel space of the camera images via perspective projection to sample the corresponding depth and uncertainty values:
\begin{equation}
\label{perspective_projection}
u_n=\frac{\tilde{u}_n}{\tilde{w}_n}, \quad v_n=\frac{\tilde{v}_n}{\tilde{w}_n}.
\end{equation}
During this process, each canonical query point, together with the LiDAR center, defines a LiDAR ray with a specific direction in 3D space. 
For each camera image, this ray intersects at most one pixel.
We project only valid query points that fall within the cameras' fields of view and image boundaries. 
Query points that cannot be projected onto any camera image are assigned null values. 
In cases of overlap or occlusion, we retain the depth and uncertainty values according to a nearest-depth criterion.
Similarly, this algorithm can also be used to sample color information in the original RGB space. 
After min-max normalization, we obtain a LiDAR-aligned relative depth map $I_D$, an uncertainty map $I_U$, and an RGB image $I_C$.

\subsubsection{Geometry-Aware Condition Encoding with Uncertainty Modulation}
\begin{figure}[t]
  \centering
  \includegraphics[width=\linewidth]{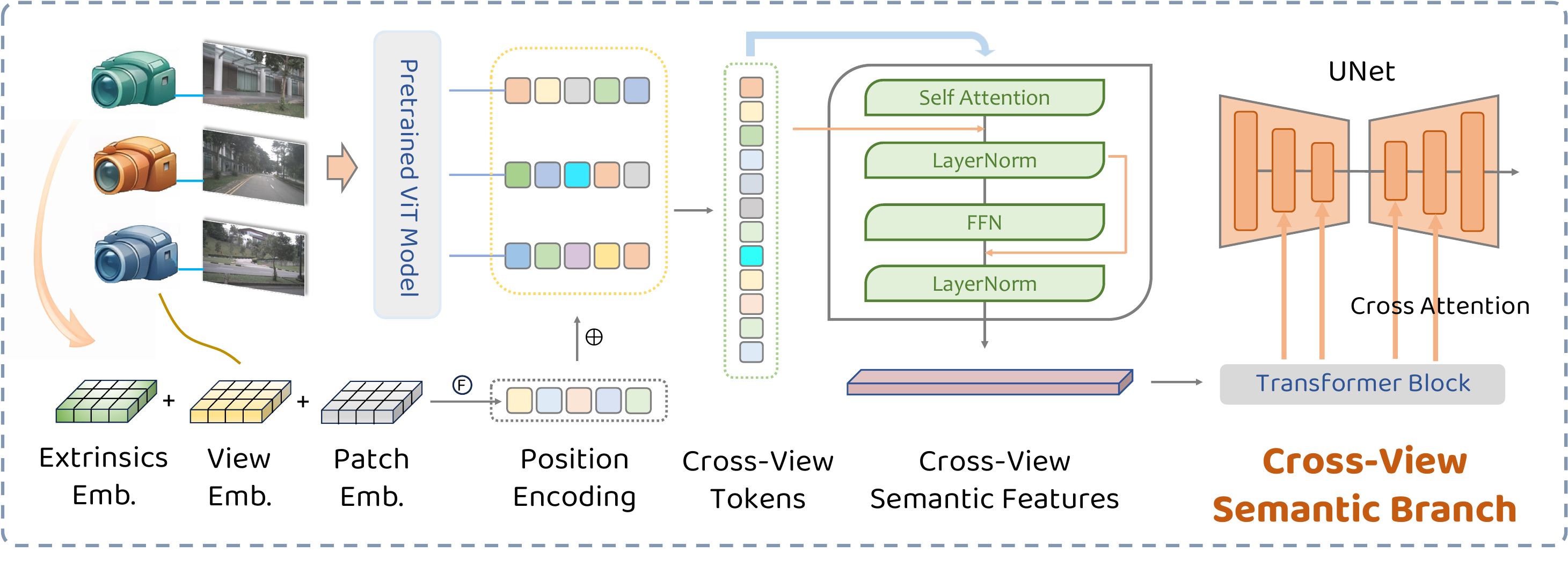}\\
  \caption{\label{semantic_branch} 
    \textbf{Cross-View Semantic Branch} first aggregates multi-view camera tokens augmented with position encodings, then extracts cross-view features through a semantic encoder with a self-attention mechanism, and finally injects them into the latent diffusion process via Transformer blocks.
}
\end{figure}
To effectively exploit these LiDAR-aligned visual priors, we design a geometry-aware conditional encoder.
Specifically, we first introduce a 2D coordinate encoding for the range-view space, and concatenate it with the depth map and RGB image as the input to the encoder.
Next, we use an encoder $\mathcal{E}_G$ to jointly model the input features. 
This encoder consists of an initial convolutional layer, several residual blocks, and downsampling layers, which progressively extract local geometric and appearance information and map them to appropriate feature scales.
Notably, rather than simply treating the uncertainty map $I_U$ as an additional input channel, we inject it into the encoding process as an explicit modulation signal through FiLM:
\begin{equation}
\tilde{H}_l = \gamma_l(I_U) \odot H_l + \beta_l(I_U),
\end{equation}
where $H_l$ denotes the output feature of the $l$-th residual block, and $\gamma_l$ and $\beta_l$ are the scaling and shifting terms generated by the FiLM module, respectively.
This mechanism enables the model to dynamically adjust the expressive strength of the conditional features according to the confidence of local depth predictions. 
As a result, the encoder can preserve explicit geometric structure while suppressing the propagation of unreliable conditional information in low-confidence regions.

This branch provides local, structured, and reliable geometric constraints for LiDAR scene generation. 
Compared with directly using raw projection results, this design more effectively integrates depth, appearance, and spatial positional information, while enhancing the consistency between the cross-modal conditions and the target LiDAR scene.

\subsection{Cross-View Semantic Conditioning}

While the geometry-aware branch provides reliable geometric constraints, it is limited in capturing long-range dependencies and cross-view consistency. 
To address this, we introduce a cross-view semantic branch to provide scene-level semantics and contextual information.
Specifically, we design a cross-view semantic encoder $\mathcal{E}_S$, which takes the flattened tokens from all views as input.
We further introduce three types of positional encodings to enhance its awareness of patch coordinates, camera identity, and camera extrinsic parameters, thereby obtaining enriched feature representations:
\begin{equation}
\tilde{X}_S = X_S + E_{patch} + E_{view} + E_{ext}.
\end{equation}
Here, \(E_{patch}\) denotes a set of learnable parameters of size \(NL \times D\), where \(N\) is the number of cameras, \(L\) is the number of image patches per camera, and \(D\) is the feature dimension.
\(E_{view}\) is a learnable embedding vector associated with each camera index.
\(E_{ext}\) is obtained by flattening the extrinsic matrix of each camera into a 16-dimensional vector and passing it through a two-layer MLP.
We then feed it into a cross-view interaction module based on multi-head attention. 
Through self-attention, this module establishes global dependencies among tokens from all views, allowing semantic information from different camera perspectives to interact within a unified feature space:
\begin{equation}
H_S=\mathrm{LayerNorm}(\tilde{X}_S + \mathrm{SelfAttn}(\tilde{X}_S)).
\end{equation}
Subsequently, $H_S$ is passed through a feed-forward network and then injected into the Transformer blocks of the UNet, where it interacts with the backbone features via cross-attention.
This branch further captures long-range scene relationships and cross-view consistency, thereby improving the plausibility of the generated results in terms of both global structure and semantic coherence.

\subsection{Dual-Branch Balanced Classifier-Free Guidance}
In the previous two sections, we introduced the geometric conditional branch and the cross-view semantic branch, respectively. 
However, in practice, the model may become biased toward one conditional branch during training, resulting in under-utilization of the other branch.

To mitigate this issue, we construct a dual-branch balanced classifier-free guidance (CFG) module. 
Standard CFG typically enhances conditional guidance at inference time by randomly dropping all conditions, thereby allowing the model to learn both conditional and unconditional distributions. 
In contrast, we extend the condition dropout mechanism beyond the conventional “all-or-nothing” scheme by additionally introducing cases where only geometric or only semantic conditions are retained. 
This enables the model not only to possess the unconditional modeling capability, but also to explicitly learn the individual effects of the two types of conditions as well as their joint contributions.
During inference, we adopt the same conditional guidance formulation as in standard CFG:
\begin{equation}
\epsilon = \epsilon_{\mathrm{uncond}} + w \cdot (\epsilon_{\mathrm{cond}} - \epsilon_{\mathrm{uncond}}),
\end{equation}
where $w$ is the guidance scale, $\epsilon_{\mathrm{uncond}}$ and $\epsilon_{\mathrm{cond}}$ denote the noise predictions under unconditional and conditional guidance, respectively.

This strategy preserves the advantages of standard CFG while further balancing the learning process between the geometric and semantic branches, enabling the model to leverage global semantic information to generate more plausible details when the depth information from the geometric branch is unreliable.
With all the above designs, our model is able to jointly leverage local geometric conditions and global semantic conditions, thereby establishing reliable cross-modal correspondences between multi-view images and LiDAR scenes and generating high-fidelity LiDAR scenes.

\begin{figure}[t]
  \centering
  \includegraphics[width=\linewidth]{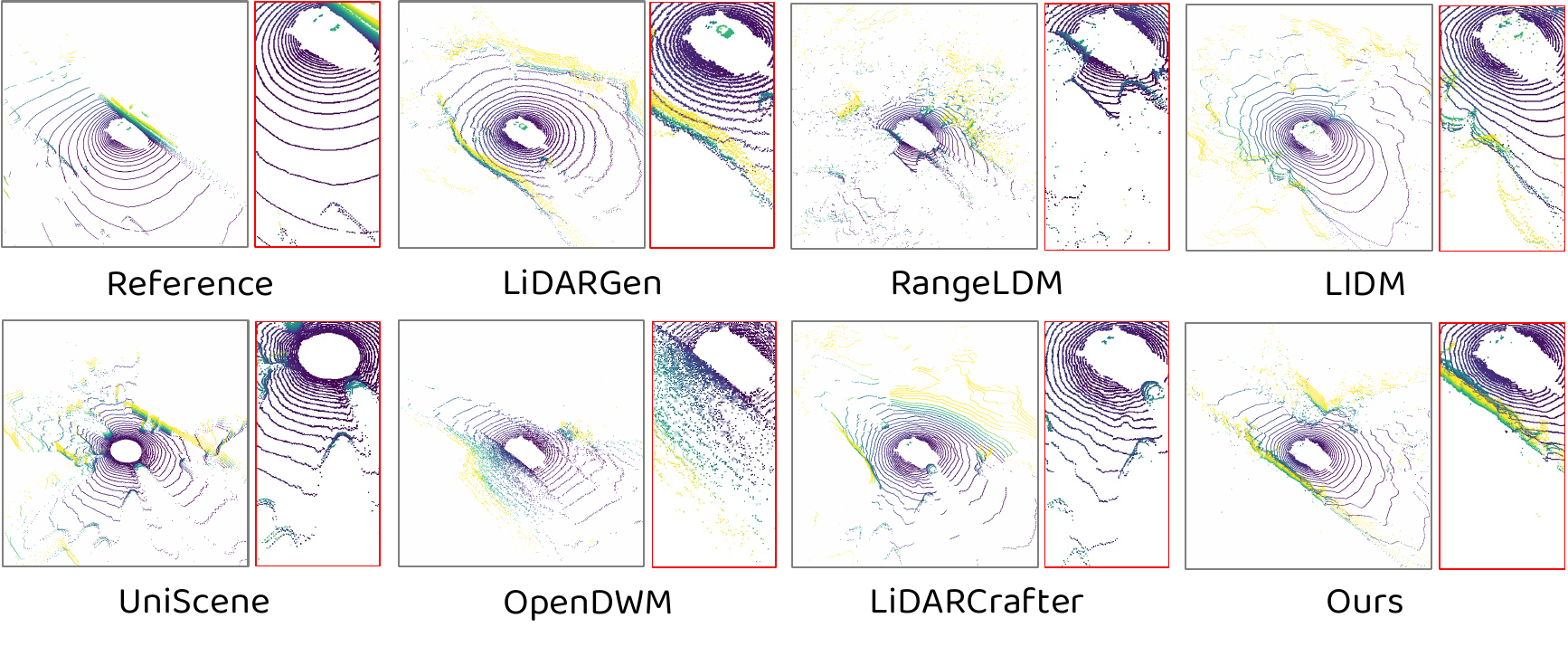}\\
  \caption{\label{gen_resluts} 
    Visualization results of LiFi and several baseline methods for LiDAR scene generation. It can be observed that our method generates details that are closest to the real LiDAR point cloud data.
}
\end{figure}


\section{Experiments}

\begin{figure}[t]
  \centering
  \includegraphics[width=\linewidth]{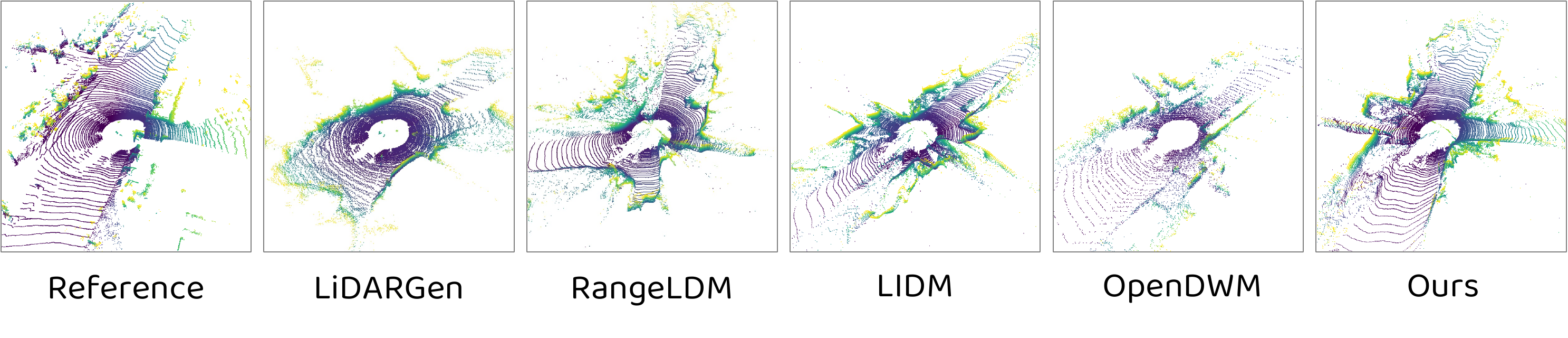}\\
  \vspace{-3mm}
  \caption{\label{kitti_results} 
    Qualitative results of LiFi and several major baseline models on the KITTI-360 dataset.
}
\vspace{-2mm}
\end{figure}

\subsection{Experimental Setups}

\noindent\textbf{Dataset.}
We conduct experiments on the challenging nuScenes dataset \cite{caesar2020nuscenes}. 
It contains over a thousand autonomous driving sequences collected in two geographically distinct urban environments with complex traffic conditions. 
The nuScenes training set includes 297,737 LiDAR scans, while the test set contains 52,423 LiDAR scans, each captured by a 32-beam LiDAR sensor. 
This dataset is widely used in various autonomous driving tasks, such as 3D object detection \cite{huang2021bevdet, li2023bevdepth, li2023end, li2024bevformer}, multi-object tracking \cite{hu2022monocular, pang2022simpletrack, zhang2022mutr3d}, trajectory prediction \cite{gu2023vip3d, liang2020pnpnet}, and end-to-end autonomous driving \cite{hu2022st, hu2023planning, jiang2023vad}.
We also conduct experiments on the KITTI-360 dataset \cite{Geiger2013IJRR}, as detailed in the supplementary material.

\noindent\textbf{Baselines.}
We compare LiFi with multiple categories of baseline methods. 
The first category includes conventional LiDAR data generation methods, covering both unconditional and conditional generation models. 
The second category consists of baselines capable of generating LiDAR from camera images, including LiDM \cite{ran2024towards} and X-Drive \cite{xie2024x}. 
The third category includes models for generating continuous LiDAR scenes. 
In addition, we apply LiFi to a downstream 3D object detection task and compare the performance of models with and without LiDAR input.

\noindent\textbf{Implementation Details.}
We conduct distributed training on four NVIDIA H200 GPUs. 
Our model is trained for 200 epochs, with a total of 176,000 optimization steps, and the learning rate is set to $1 \times 10^{-4}$. 
We use the DDIM sampler \cite{song2020denoising} for denoising, and set the total number of diffusion timesteps $T$ to 1024.
The UNet consists of three downsampling blocks and three upsampling blocks, each containing two Transformer blocks. 
We employ DepthAnything3 \cite{lin2025depth} as the pretrained visual encoder, which can predict spatially consistent geometry from an arbitrary number of input views.
It is an instantiation choice that matches the requirements, rather than an irreplaceable component of the framework.
During inference, we perform 50 denoising steps to generate the target range image. 
More experimental settings and implementation details are provided in the supplementary material.

\subsection{LiDAR Data Generation}
This subsection evaluates the performance of LiFi on the task of generating high-fidelity LiDAR point clouds. 
We compare our method against both unconditional and conditional generation baselines. 
Following the evaluation protocol of LiDARGen \cite{LiDARGen2022}, we adopt MMD, JSD, and two perception-based Fréchet distances as evaluation metrics. 
As shown in Table~\ref{table_generation}, our method achieved clear improvements across multiple metrics.

\begin{table}[h]
\centering
\caption{The main quantitative results of LiFi and several baseline methods on the nuScenes dataset. The MMD values are reported in $10^{-5}$ and JSD in $10^{-2}$.}
\label{table_generation}
\setlength{\tabcolsep}{3.5pt}
\renewcommand{\arraystretch}{1.1}
\resizebox{0.65\columnwidth}{!}{
\small
\begin{tabular}{c|l|ccccc} 
\toprule
& \textbf{Method} & \textbf{Venue} & \textbf{MMD $\downarrow$} & \textbf{JSD $\downarrow$} & \textbf{FRD $\downarrow$} & \textbf{FPD $\downarrow$} \\
\midrule \midrule 
\multirow{3}{*}{\rotatebox{90}{Uncond.}} 
& LiDARGen \cite{LiDARGen2022}   & ECCV'22 & 55.14 & 27.69 & -- & -- \\
& RangeLDM \cite{hu2024rangeldm}   & ECCV'24 & 72.01 & 8.43 & 1172.06 & 441.16 \\
& LiDM \cite{ran2024towards}       & CVPR'24 & 4.57 & 10.61 & 2358.15 & 420.36 \\
\midrule 
\multirow{5}{*}{\rotatebox{90}{Cond.}} 
& UniScene \cite{li2025uniscene}   & CVPR'25 & 107.18 & 12.35 & 1636.60 & 230.89 \\
& OpenDWM \cite{ni2025maskgwm}    & CVPR'25 & 10.75 & 7.69 & 2607.81 & 146.30 \\
& LiDARCrafter \cite{liang2025lidarcrafter} & AAAI'26 & 24.69 & 7.37 & 1183.11 & \textbf{36.17} \\
& \cellcolor{myyellow}\textbf{LiFi} &\cellcolor{myyellow} -- & \cellcolor{myyellow}\textbf{2.78} & \cellcolor{myyellow}\textbf{3.66} & \cellcolor{myyellow}\textbf{906.71} & \cellcolor{myyellow}126.31 \\
\bottomrule
\end{tabular}
}
\end{table}

\begin{figure}[t]
  \centering
  \includegraphics[width=0.88\linewidth]{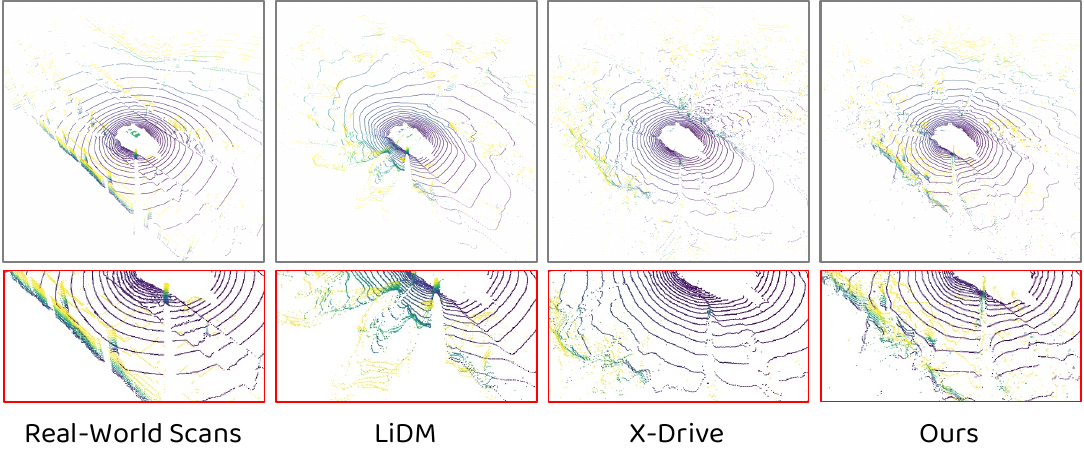}\\
  \caption{\label{results_c2l} 
    Qualitative results of LiDAR scene generation from camera images. The first row shows the full scenes, and the second row presents zoomed-in local details.
}
\end{figure}

\begin{table}[ht]
\centering
\caption{The main quantitative results of LiFi and several baseline methods on the KITTI-360 dataset. The MMD values are reported in $10^{-4}$ and JSD in $10^{-2}$.}
\label{table_kitti_generation}
\setlength{\tabcolsep}{3.5pt}
\renewcommand{\arraystretch}{1.1}
\resizebox{0.62\columnwidth}{!}{
\small
\begin{tabular}{l|c|cccc} 
\toprule
\textbf{Method} & \textbf{Venue} & \textbf{MMD $\downarrow$} & \textbf{JSD $\downarrow$} & \textbf{FRD $\downarrow$} & \textbf{FPD $\downarrow$} \\
\midrule \midrule 
LiDARGen \cite{LiDARGen2022} & ECCV'22 & 13.36 & 13.67 & 2415.23 & 102.80 \\
RangeLDM \cite{hu2024rangeldm} & ECCV'24 & 4.31 & 9.42 & 1362.25 & 68.22 \\
LiDM \cite{ran2024towards} & CVPR'24 & 9.31 & 19.00 & 2894.65 & 253.91 \\
OpenDWM \cite{ni2025maskgwm} & CVPR'25 & 12.93 & 14.63 & 1966.36 & 436.17 \\
\rowcolor{myyellow}%
\textbf{LiFi} & -- &\textbf{1.99} & \textbf{8.77} & \textbf{1107.21} & \textbf{21.40} \\
\bottomrule
\end{tabular}
}
\end{table}

\subsection{Camera-to-LiDAR Generation}

This subsection compares LiFi with several baselines capable of generating LiDAR point clouds from camera images. 
We additionally introduce the CD and EMD metrics, which enable a more rigorous evaluation of point-level fidelity in the generated scenes, to assess the model’s performance in generating geometrically consistent LiDAR scenes.
We also introduce the DAS metric \cite{xie2024x} to evaluate the cross-modal consistency.
As shown in Table~\ref{table_c2l_generation}, LiFi achieves substantially better point-level fidelity and cross-modal consistency than existing camera-to-LiDAR baselines.

We further evaluate the quality of LiDAR data generated by LiFi across different depth ranges to investigate how its generation performance varies with scene distance and to analyze the associated failure modes. 
Detailed results are provided in the supplementary material.

\begin{table}[ht]
\centering
\caption{Quantitative results on the nuScenes dataset. These baseline models are used for generating LiDAR scenes from camera images. The MMD values are reported in $10^{-5}$ and JSD in $10^{-2}$. Inference time is evaluated in s/frame. Evaluation is conducted on a single NVIDIA H200 GPU.}
\label{table_c2l_generation}
\setlength{\tabcolsep}{3.5pt}
\renewcommand{\arraystretch}{1.1}
\resizebox{0.76\columnwidth}{!}{
\small
\begin{tabular}{l|ccccccc} 
\toprule
\textbf{Method} & \textbf{Venue} & \textbf{MMD $\downarrow$} & \textbf{JSD $\downarrow$} & \textbf{CD $\downarrow$} & \textbf{EMD $\downarrow$} & \textbf{DAS $\downarrow$} & \textbf{Time}\\
\midrule \midrule 
LiDM \cite{ran2024towards} & CVPR'24 & 8.02 & 9.97 & 4.57 & 99.71 & 2.17 & \textbf{0.34}\\
X-Drive \cite{xie2024x}    & ICLR'25 & 48.58 & 10.54 & 4.73 & 179.62 & 1.52 & 1.80\\
\rowcolor{myyellow}%
\textbf{LiFi} & -- & \textbf{2.78} & \textbf{3.66} & \textbf{0.83} & \textbf{50.08} & \textbf{1.02} & 0.52\\
\bottomrule
\end{tabular}
}
\end{table}

\subsection{Temporal Consistency of Continuous Scene Generation }
We further evaluate the temporal consistency of LiDAR scenes generated by LiFi under consecutive camera inputs.
While this approach does not explicitly incorporate mechanisms for temporal generation, inputting consecutive camera frames nevertheless produces LiDAR sequences that maintain good temporal consistency.
Following LiDARCrafter~\cite{liang2025lidarcrafter}, we evaluate these methods using CTC and TTCE, two metrics designed to measure temporal consistency.
The results are shown in Table~\ref{table_temporal_consistency}.

\begin{table}[ht]
\caption{Quantitative results of LiFi for continuous LiDAR scene generation. Here, TTCE and CTC are adopted to evaluate the temporal consistency of the generated results.}
\label{table_temporal_consistency}
\setlength{\tabcolsep}{3.5pt}
\renewcommand{\arraystretch}{1.11}
\centering
\resizebox{0.7\columnwidth}{!}{
\small 
\begin{tabular}{@{}l|c|cc|ccccc@{}}
\toprule
\multirow{2}{*}{\textbf{Method}} & \multirow{2}{*}{\textbf{Venue}} & \multicolumn{2}{c|}{\textbf{TTCE}} & \multicolumn{5}{c}{\textbf{CTC}$\downarrow$} \\
 &  & \textbf{3} & \textbf{4} & \textbf{1} & \textbf{2} & \textbf{3} & \textbf{4} & \textbf{10}\\ 
\midrule \midrule
UniScene \cite{li2025uniscene} & CVPR'25 & 2.74 & 3.69 & 0.90 & 1.84 & 3.64 & 3.90 & -- \\
OpenDWM \cite{ni2025maskgwm} & CVPR'25 & 2.68 & 3.65 & 1.02 & 2.02 & 3.37 & 5.05 & -- \\
OpenDWM-DiT \cite{ni2025maskgwm} & CVPR'25 & 2.71 & 3.66 & 0.89 & 1.79 & 3.06 & 4.64 & -- \\ 
LiDARCrafter \cite{liang2025lidarcrafter} & AAAI'26 & \textbf{2.65} & \textbf{3.56} & 1.12 & 2.38 & 3.02 & 4.81 & --  \\ 
\rowcolor{myyellow}%
\textbf{LiFi} & \textbf{--} & 2.67 & 3.61 & \textbf{0.80} & \textbf{1.11} &  \textbf{1.52} & \textbf{2.22} & \textbf{3.19} \\ 
\bottomrule
\end{tabular}
}
\end{table}

\begin{wraptable}[11]{r}{0.50\columnwidth}
\vspace{-2.0\baselineskip}
\centering
\caption{Quantitative results of applying LiFi to the downstream 3D object detection task. We compare the performance of the camera-only model and the fusion-based model.}
\label{table_downstream}
\vspace{2mm}
\setlength{\tabcolsep}{3.5pt}
\renewcommand{\arraystretch}{1.1}
\small
\resizebox{0.50\columnwidth}{!}{
\begin{tabular}{l|cc}
\toprule
\textbf{Method} & \textbf{mAP $\uparrow$} & \textbf{NDS $\uparrow$} \\
\midrule \midrule
BEVDet (Camera Only) \cite{huang2021bevdet} & 0.278 & 0.341 \\
BEVFusion (+Real LiDAR) \cite{liang2022bevfusion} & 0.529 & 0.533 \\
\rowcolor{myyellow}%
\textbf{BEVFusion (+Gen LiDAR)} \cite{liang2022bevfusion} & \textbf{0.526} & \textbf{0.520} \\
\bottomrule
\end{tabular}
}
\vspace{-1mm}
\end{wraptable}

\subsection{Downstream Task: 3D Object Detection}
To assess the practical value of LiFi in real-world applications, we apply it to the downstream 3D object detection task in autonomous driving.
We retrained the detector using the generated LiDAR data and compared the camera-only (BEVDet \cite{huang2021bevdet}) and camera-LiDAR fusion baselines (BEVFusion \cite{liang2022bevfusion}).
The experimental results in Table~\ref{table_downstream} demonstrate that LiDAR point clouds generated by LiFi have certain value in downstream tasks of autonomous driving.

\subsection{Ablation Study}

In this subsection, we conduct ablation studies to evaluate the contribution of each component in the LiFi framework. 
We first conduct ablation studies on the two branches. 
We further perform ablations on the depth map, the original RGB image, and the uncertainty modulation mechanism in the geometric branch. 
The results are shown in Table~\ref{table_ablation_main}.
Finally, we perform ablation studies on the CFG module, including the dual-branch balanced design and the guidance scale parameter. 
The corresponding results are presented in Table~\ref{table_ablation_cfg}.
Furthermore, we conduct ablation studies on the number of input image views, the resolution of input range images, and the number of denoising steps during inference. 
Detailed results are provided in the supplementary material.

\begin{table}[h]
\centering
\setlength{\tabcolsep}{3.5pt}
\renewcommand{\arraystretch}{1.0}

\begin{minipage}[t]{0.48\columnwidth}
\centering
\caption{Ablation results on the two branches and the main conditions in the LiFi framework. The MMD is reported in $10^{-5}$ and JSD in $10^{-2}$.}
\label{table_ablation_main}
\small
\begin{tabular}{l|cc} 
\toprule
\textbf{LiFi}  & \textbf{MMD $\downarrow$} & \textbf{JSD $\downarrow$} \\
\midrule \midrule 
w/o Geometry-Aware Branch   & 5.20 & 4.27 \\
w/o Cross-View Semantic Branch   & 3.57 & 3.67 \\
\midrule
w/o Depth Map Condition  & 3.57 & 4.12 \\
w/o RGB Image Condition  & 2.80 & 3.92 \\
w/o Uncertainty Modulation   & 2.87 & 3.80 \\
\midrule
\rowcolor{myyellow}%
\textbf{w/ All} & \textbf{2.78} & \textbf{3.66} \\
\bottomrule
\end{tabular}
\end{minipage}
\hfill
\begin{minipage}[t]{0.48\columnwidth}
\centering
\caption{Ablation results of the dual-branch balanced CFG module.
$w$ is the CFG scale. The MMD is reported in $10^{-5}$ and JSD in $10^{-2}$.}
\label{table_ablation_cfg}
\small
\begin{tabular}{l|cc} 
\toprule
\textbf{LiFi}  & \textbf{MMD $\downarrow$} & \textbf{JSD $\downarrow$} \\
\midrule \midrule 
w/o Classifier-Free Guidance  & 6.32 & 4.85 \\
w/o Dual-Branch Balanced  & 3.02 & 4.31 \\
\midrule
Guidance scale $w = 2.0$  & 5.71 & 3.76 \\
Guidance scale $w = 4.0$  & 3.45 & 3.81 \\
Guidance scale $w = 8.0$  & \textbf{2.37} & 4.13 \\
\midrule
\rowcolor{myyellow}%
\textbf{w/ All $(w = 6.0)$} & 2.78 & \textbf{3.66} \\
\bottomrule
\end{tabular}
\end{minipage}
\end{table}


\section{Conclusion}

We propose LiFi, a dedicated framework for generating realistic LiDAR scenes from multi-view camera images. 
We introduce a dual-branch conditioning strategy to guide the latent diffusion process, effectively addressing the challenges of cross-modal geometric alignment and cross-view semantic aggregation. 
Experimental results show that LiFi is capable of generating high-fidelity LiDAR scenes.
Moreover, the LiDAR point cloud generated by it has certain value for some downstream tasks of autonomous driving.
In the future, we expect LiFi to play an important role in applications such as simulation, planning, and control in various complex environments and long-tail scenarios.


\begin{ack}
This research was supported by the Scientific Research Innovation Capability Support Project for Young Faculty (U40) of the Ministry of Education of China (SRICSPYF-ZY2025019), the National Natural Science Foundation of China (No.62272433), Anhui Provincial Natural Science Foundation (No.2508085ZD011), Key Science \& Technology Project of Anhui Province (No. 202523o09050004) and the Fundamental Research Funds for the Central Universities.
\end{ack}


\bibliographystyle{ref}
\small
\bibliography{ref}
\normalsize

\clearpage

\appendix

\section{Overview of the Appendix}
The appendix is structured into the following sections to supplement the main paper.
The first part provides a discussion on the limitations of this research and directions for future work.
The second part presents additional experimental details.
The third part provides an in-depth theoretical analysis of the proposed modules.
The fourth part supplements the experimental results.

\section{Limitations and Future Work}
Our work presents several limitations. 
Primarily, relying on the range image representation is not entirely lossless, resulting in the omission of certain camera image details within the synthesized LiDAR point clouds. 
Refining this representation in future research holds the potential to further boost model performance.
Furthermore, the overall architecture of our method is not specifically designed for explicit temporal modeling. 
Although this work does not focus on generating temporal LiDAR sequences, incorporating temporal consistency mechanisms in future work could extend the applicability of our method to a broader range of scenarios.

\section{More Implementation Details}
During training, our model is trained for 200 epochs with a total of 176,000 optimization steps. 
The total batch size is set to 32, and the learning rate is set to \(1 \times 10^{-4}\), with a cosine annealing scheduler applied. 
We perform distributed training on 4 NVIDIA H200 GPUs, with a total training time of approximately 20 hours. 
For denoising, we employ the DDIM sampler \cite{song2020denoising}, with the total number of diffusion timesteps  \(T\) set to 1024.
During inference, we perform 50 denoising steps to generate point clouds. 
The inference is conducted on a single NVIDIA H200 GPU.

We adopt DepthAnything3~\cite{lin2025depth} as the pre-trained visual encoder. 
For the semantic branch, we follow DepthAnything3 and employ DINOv2 \cite{oquab2023dinov2} to extract visual features. 
Based on the nuScenes dataset, we use six camera images from different viewpoints as input to generate each scene.

In the geometry-aware branch, the encoder primarily consists of multiple convolutional layers, normalization layers, downsampling layers, zero-convolution layers, and FiLM modules. 
In the cross-view semantic branch, the encoder mainly includes a multi-head self-attention layer, a feed-forward network (composed of two fully connected layers), and two Layer Normalization layers. 
Both the attention layer and the FFN employ residual connections.

The VAE in the LDM is initialized with pre-trained weights from RangeLDM \cite{hu2024rangeldm}. 
The range image resolution is set to [1024, 32], and the VAE uses a downsampling factor of 4. 
The UNet consists of three downsampling blocks (the latter two incorporating Transformer layers) and three upsampling blocks (the first two incorporating Transformer layers). 
We employ circular convolution instead of standard convolution in both the VAE and the LDM.

\section{Detailed Description of Evaluation Metrics}
In the main text, we introduced the Maximum Mean Discrepancy (MMD), Jensen-Shannon Divergence (JSD), Fréchet Point cloud Distance (FPD), and Fréchet Range Image Distance (FRD) metrics to evaluate the performance of the generator. 
Below, we provide a detailed explanation.

Maximum Mean Discrepancy (MMD) is a non-parametric method for measuring the distance between two probability distributions. 
The fundamental idea is to compute the distance between the kernel means of two sample sets in a high-dimensional space; by comparing the means mapped to this space, one can determine whether the sample distributions are consistent:
\begin{equation}
\label{mmd}
    MMD = \frac{1}{N^2} \sum_{i}^{N} \sum_{j}^{N} k(p_i, p_j) + \frac{1}{N^2} \sum_{i}^{N} \sum_{j}^{N} k(q_i, q_j) - \frac{2}{N^2} \sum_{i}^{N} \sum_{j}^{N} k(p_i, q_j).
\end{equation}

Jensen-Shannon Divergence (JSD) is a symmetric information-theoretic measure for quantifying the similarity between two probability distributions, derived from the Kullback-Leibler (KL) divergence and commonly used for distribution comparison. 
The essence of JSD is to evaluate the information loss of each distribution with respect to their uniform mixture:
\begin{equation}
\label{jsd}
JSD = \frac{1}{2}\sum_{i}^{N} (p_i \log\frac{p_i}{m_i} + q_i \log\frac{q_i}{m_i}).
\end{equation}
We compute the MMD and JSD using 160 × 100 2D histograms on the bird’s-eye view (BEV) plane.

Fréchet Point cloud Distance (FPD) is used to measure the distance between the generated point cloud distribution and the real point cloud distribution. 
It utilizes a pre-trained point cloud feature extraction network (such as PointNet or PointNet++) to extract global features, and then computes the Fréchet distance between two Gaussian distributions:
\begin{equation}
\label{fpd}
FPD=\|\mu_r - \mu_g\|^2_2 + \text{Tr}(\sigma_r + \sigma_g - 2 \sqrt{\sigma_r \cdot \sigma_g}).
\end{equation}

Fréchet Range Image Distance (FRD) is a variant specifically designed for LiDAR data. 
It first converts the point cloud into a Range Image and then computes the Fréchet distance. It uses a 2D CNN (such as ResNet) to extract features. 
The formula is identical to that of FPD, but the source of the features differs.

To evaluate the temporal consistency of the synthesized sequences, we follow LiDARCrafter~\cite{liang2025lidarcrafter} and adopt the TTCE (Translation and Transformation Consistency Error) and CTC (Chamfer Temporal Consistency) metrics.

Let $T_{t}^{gt}=[R_{t}^{gt} \mid t_{t}^{gt}]$ denote the ground-truth transformation from frame $t$ to $t+1$, where $R_{t}^{gt}$ represents the rotation matrix and $t_{t}^{gt}$ is the translation vector. 
Let $T_{t}^{pred}=[R_{t}^{pred} \mid t_{t}^{pred}]$ be the transformation estimated from the generated point clouds via Iterative Closest Point (ICP). 
The TTCE is calculated as follows:
\begin{equation}
TTCE = \frac{1}{T-1} \sum_{t=1}^{T-1} \| t_{t}^{pred} - t_{t}^{gt} \|_F, 
\end{equation}
where $\|\cdot\|_F$ denotes the Frobenius norm and $T$ is the total number of frames in the sequence.

Furthermore, let $\mathcal{P}_t$ and $\mathcal{P}_{t+1}$ be the generated point clouds at frames $t$ and $t+1$, respectively. Using the ground-truth transformation $T_{t}^{gt}$, we align $\mathcal{P}_{t+1}$ to the coordinate system of frame $t$:
\begin{equation}
\tilde{\mathcal{P}}_{t+1} = \left(R_{t}^{gt}\right)^{-1} \left( \mathcal{P}_{t+1} - t_{t}^{gt} \right). 
\end{equation}
Subsequently, the Chamfer Distance (CD) between $\mathcal{P}_t$ and the aligned $\tilde{\mathcal{P}}_{t+1}$ is computed:
\begin{equation}
 CD\left(\mathcal{P}_t, \tilde{\mathcal{P}}_{t+1}\right) = \frac{1}{\|\mathcal{P}_t\|} \sum_{x \in \mathcal{P}_t} \min_{y \in \tilde{\mathcal{P}}_{t+1}} \| x - y \|_2^2 + \frac{1}{\|\tilde{\mathcal{P}}_{t+1}\|} \sum_{y \in \tilde{\mathcal{P}}_{t+1}} \min_{x \in \mathcal{P}_t} \| y - x \|_2^2. 
\end{equation}
The final CTC score is defined as the average Chamfer Distance across all adjacent frame pairs:
\begin{equation}
CTC = \frac{1}{T-1} \sum_{t=1}^{T-1} CD\left(\mathcal{P}_t, \tilde{\mathcal{P}}_{t+1}\right).
\end{equation}

Moreover, we introduce Chamfer Distance (CD)~\cite{fan2017point} and Earth Mover’s Distance (EMD) to evaluate the fidelity of the generated point clouds in the camera-to-LiDAR task.

Chamfer Distance is a commonly used metric for measuring similarity between point clouds. 
It quantifies the “alignment” distance between two sets of points, representing the sum of the distances from each point in one set to its nearest neighbor in the other set. 
By aggregating these nearest neighbor distances across all points, Chamfer Distance captures the overall distribution difference:
\begin{equation}
\label{Chamfer_Distance}
    CD(P^{s}_{\text{gen}}, P^{s}_{\text{gt}})= \frac{1}{M} \sum_{x \in P^{s}_{\text{gt}}} \min_{y \in P^{s}_{\text{gen}}} \|x - y\|_2^2 + \frac{1}{N} \sum_{y \in P^{s}_{\text{gen}}} \min_{x \in P^{s}_{\text{gt}}} \|y - x\|_2^2.
\end{equation}

Earth Mover’s Distance (EMD) measures the minimum average cost required to transport the generated point cloud to the real point cloud. 
Compared with CD, EMD better reflects differences in the global structure and density distribution between two point clouds:
\begin{equation}
\label{EMD}
EMD(P,Q)=
\min_{T \geq 0}
\sum_{i=1}^{N}
\sum_{j=1}^{M}
T_{ij}\, d(p_i,q_j),
\quad
\text{s.t. }
\sum_{j=1}^{M}T_{ij}=a_i,\ 
\sum_{i=1}^{N}T_{ij}=b_j.
\end{equation}

Considering the actual requirements of autonomous driving scenarios, we measured the above indicators within the range of [-50.0, 50.0] m.

In downstream detection tasks for autonomous driving, we introduce the mean Average Precision (mAP) and nuScenes Detection Score (NDS) metrics to evaluate detection accuracy and overall perception quality.

The mAP metric measures the precision-recall performance of the detection model across different categories:
\begin{equation}
mAP = \frac{1}{C} \sum_{c=1}^{C} AP_c,
\end{equation}
where \( AP \) is the area under the Precision-Recall curve:
\begin{equation}
AP = \sum_n (R_n - R_{n-1}) \cdot P_{interp}(R_n),
\end{equation}
where \( P \) and \( R \) represent precision and recall, respectively.

The NDS summarizes overall 3D detection performance:
\begin{equation}
NDS = \frac{1}{10} \left(5 \cdot mAP + \sum_{k \in {ATE, ASE, AOE, AVE, AAE}} (1 - \min(1, e_k)) \right),
\end{equation}
where \( ATE \), \( ASE \), \( AOE \), \( AVE \), and \( AAE \) represent position error, size error, orientation error, velocity error, and attribute error, respectively.

\section{Standard Range-View Projection and Corrected Version}
Range images offer a compact representation of LiDAR data by parametrizing unstructured point clouds in 3D space into a 2D pixel space. 
This projection process is reversible, and after applying it, both training and inference costs for the model can be significantly reduced. Consequently, representing LiDAR data with range images has become one of the prevailing trends in the field.

The standard range image mapping method projects each point \( p = (x, y, z) \in \mathbb{R}^3 \) in the point cloud into the spherical coordinates \( (r, \theta, \phi) \) of the range image using spherical coordinate projection:
\begin{equation}
\label{range_view}
    r = \sqrt{x^{2} + y^{2} + z^{2}}, \quad
    \theta = \arctan(y, x), \quad
    \phi = \arctan\left(z, \sqrt{x^{2} + y^{2} + z^{2}}\right),
\end{equation}
where \( r \) is the depth value, \( \theta \) is the azimuth angle, and \( \phi \) is the elevation angle.
RangeLDM \cite{hu2024rangeldm} points out that multiple lasers in a LiDAR system do not share the same origin during measurement, which can introduce errors during coordinate transformation and lead to incorrect range-view data distributions. 
Therefore, we adopt its corrected range image representation as follows:
\begin{equation}
\label{corrected_range_view}
    r = \sqrt{x^{2} + y^{2} + (z - h_{j})^{2}}, \quad
    \theta = \arctan(y, x), \quad
    \phi = \phi_{j},
\end{equation}
where \( h_{j} \) and \( \phi_{j} \) represent the sensor height and elevation angle estimated by Hough voting, respectively.
In subsequent computations, both projection methods are processed in exactly the same way, including converting spherical coordinates \( (r, \theta, \phi) \) to range image pixel coordinates \( (u, v) \):
\begin{equation}
    u = W - ((\theta + \pi) / 2\pi) \cdot W, \quad
    v = H - j,
\end{equation}
as well as normalizing the depth values of the point cloud to obtain the range image \( I \).

In the subsequent calculation of the CD metric, it is necessary to reverse-convert the range image back into a point cloud. 
In the corrected version, the standard LiDAR's fifth dimension, which is the beam index \( j \) for each point, is directly used during the forward projection process.
However, the generated point clouds do not contain the beam index dimension.
Therefore, we introduce the following formula to compute the beam index for each point:
\begin{equation}
\begin{aligned}
    j = (\text{Rad2Deg}(\arcsin(\frac{z}{d})) - fov_{\text{down}}) / (fov_{\text{up}} - fov_{\text{down}}) \cdot (H - 1),
\end{aligned}
\end{equation}
where \( fov_{\text{up}} \) and \( fov_{\text{down}} \) refer to the elevation angle range of the original LiDAR acquisition.

\section{Theoretical Analysis of the Geometry-Aware Mechanism}

\subsection{Computational Complexity}
Let the resolution of the high-resolution camera depth map be \( H_{img} \times W_{img} \), and the target LiDAR range image resolution be \( H_r \times W_r \).
Conventional projection methods require iterating over each pixel of the depth map, calculating the 3D coordinates, and projecting them onto a cylindrical coordinate system, followed by rasterization. The computational complexity of this process is \( \mathcal{O}(H_{img} \cdot W_{img}) \).
Since this process involves mapping from a high-resolution image to a low-resolution LiDAR grid, it typically results in many-to-one write conflicts. 
To address the occlusion problem, additional aggregation operations are required, which introduce extra computational overhead.

In contrast, the reverse sampling method starts directly from the target LiDAR grid. For each query point \( q_k \), the projection via the camera is:
\begin{equation}
[\tilde{u}_n, \tilde{v}_n, \tilde{w}_n]^T = K_n (R_n \cdot q_k + t_n),
\end{equation}
where \( K_n \), \( R_n \), and \( t_n \) are the camera's intrinsic matrix, rotation matrix, and translation vector. The overall time complexity is strictly constrained to \( \mathcal{O}(H_r \cdot W_r) \).
Since in physical scenes, \( H_r \cdot W_r \ll H_{img} \cdot W_{img} \), the reverse sampling significantly reduces the computational overhead.

\subsection{Spatial Sampling Distribution}
The forward projection can be formalized as a nonlinear mapping from the image plane \( \Omega_{\text{img}} \subset \mathbb{R}^2 \) to the three-dimensional space:
\begin{equation}
\mathcal{F}(u,v) = D(u,v) \cdot K^{-1} [u,v,1]^\top,
\end{equation}
where \( D(u,v) \) is the depth at pixel \( (u,v)\), and \( K^{-1} \) is the inverse of the camera's intrinsic matrix.
Due to the nonlinearity of perspective projection, the Jacobian determinant satisfies:
\begin{equation}
\left|\det \frac{\partial (x,y)}{\partial (X,Y)}\right| \propto \frac{1}{Z^2},
\end{equation}
where \( Z \) is the depth. Therefore, uniform sampling on the image induces a depth-dependent non-uniform distribution in 3D space:
\begin{equation}
\rho(X,Y,Z) \propto \frac{1}{Z^2}.
\end{equation}
This results in over-sampling in nearby regions and severe under-sampling in distant regions, leading to significant aliasing and ghost artifacts when mapped to the regular LiDAR range image grid.
In contrast, our proposed reverse sampling method directly constructs the set of query points \( Q_D = \{q_k\} \) in the target LiDAR domain, with a distribution that is consistent with the LiDAR's spherical scanning pattern. 
Specifically, the query points are generated via spherical coordinate parametrization, which is equivalent to performing structured discrete sampling on the target spherical domain, rather than performing non-uniform reprojection in the image domain.

For each query point \( q_k \), we project it onto the image plane to obtain continuous coordinates \( (u_n, v_n) \), and then use nearest-neighbor sampling to extract the corresponding value from the depth map:
\begin{equation}
\hat{D}(q_k) = D_n(\lfloor u_n \rceil, \lfloor v_n \rceil).
\end{equation}
This process can be viewed as a zero-order hold approximation of the continuous depth field, with the error arising from pixel quantization. 
Assuming the underlying depth field \( D^* \) is Lipschitz continuous, the sampling error satisfies:
\begin{equation}
|\hat{D}(q_k) - D^*(u_n, v_n)| \leq L \cdot \delta,
\end{equation}
where \( L \) is the local Lipschitz constant, and \( \delta \) is the pixel scale.

\subsection{Weighted Estimation and Optimization Dynamics}
In the multi-view geometric conditioning process, due to occlusion and depth estimation errors, the intermediate features \( H_l \) of the encoder are often corrupted by spatially correlated and unevenly distributed noise. 
We model this as:
\begin{equation}
H_l(x, y) = H_{\text{true}}(x, y) + \epsilon(x, y),
\end{equation}
where the noise term satisfies:
\begin{equation}
\epsilon(x, y) \sim \mathcal{N}(0, \sigma^2(x, y)),
\end{equation}
and the noise variance \( \sigma^2(x, y) \) is monotonically related to the uncertainty map \( I_U(x, y) \), i.e.,
\begin{equation}
\sigma^2(x, y) \propto I_U(x, y).
\end{equation}
This model corresponds to a typical heteroscedastic noise setting, where different spatial locations exhibit varying levels of noise intensity.

In the above setup, a natural objective is to weight the features based on local noise intensity to reduce the impact of high-noise regions on the overall estimate. 
Consider the following weighted form:
\begin{equation}
\hat{H}(x, y) = w(x, y) \cdot H_l(x, y),
\end{equation}
where the weight \( w(x, y) \) should decrease as the noise variance increases.
Classical statistical estimation theory indicates that, under certain distribution assumptions, a reasonable weight form is typically inversely proportional to the noise variance:
\begin{equation}
w(x, y) \propto \frac{1}{\sigma^2(x, y)}.
\end{equation}
In our setup, since \( \sigma^2(x, y) \) is approximately characterized by \( I_U(x, y) \), the weight can be treated as a function of \( I_U \):
\begin{equation}
w(x, y) = g(I_U(x, y)),
\end{equation}
where \( g(\cdot) \) is a monotonically decreasing function.

We introduce uncertainty-based FiLM modulation into the model:
\begin{equation}
\tilde{H}_l = \gamma_l(I_U) \odot H_l + \beta_l(I_U),
\end{equation}
where \( \gamma_l(\cdot) \) and \( \beta_l(\cdot) \) are nonlinear mappings learned by the neural network.
This can be understood as a learnable heteroscedastic weighted estimator, where the weight function is learned in a data-driven manner. 
In contrast, if the uncertainty \( I_U \) were directly used as an additional input channel to the network, the model would need to implicitly learn the above weighting relationship through high-dimensional nonlinear mappings. 
This approach presents challenges such as high learning difficulty, poor sample efficiency, and optimization instability.

It also has a direct impact on the backpropagation process. For the loss function \( \mathcal{L} \), we have:
\begin{equation}
\frac{\partial \mathcal{L}}{\partial H_l} = \gamma_l(I_U) \odot \frac{\partial \mathcal{L}}{\partial \tilde{H}_l}.
\end{equation}
As a result, the gradient magnitude is modulated by \( \gamma_l(I_U) \), and when \( I_U(x, y) \) is large, \( \gamma_l(I_U) \) tends to decrease, causing the gradient to be attenuated. 
This enables the model to adaptively suppress gradient interference from unreliable regions during optimization, thereby improving training stability.

\section{Theoretical Analysis of the Cross-View Semantic Branch}
In multi-view systems, the 2D images from different viewpoints are projections of the same scene under different camera poses, and the imaging process can be represented as a projection mapping \( \pi: \mathbb{R}^3 \to \mathbb{R}^2 \).
Due to the lack of depth information, the 3D correspondence of any pixel \((u,v)\) is a set of points along a ray:
\begin{equation}
\pi^{-1}(u, v) = \{X \mid \pi(X) = (u, v)\},
\end{equation}
which means that relying solely on 2D features for cross-view interaction is theoretically ill-posed.

To address this, we introduce patch encoding \( E_{\text{patch}} \), viewpoint encoding \( E_{\text{view}} \), and camera extrinsic parameter encoding \( E_{\text{ext}}(T) \), to construct pose-conditioned features and explicitly introduce the conditional distribution into the feature space. 
This design reduces the feature confusion caused by projection uncertainty.

Building on this, we employ a self-attention mechanism to model the cross-view dependencies. This process can be formalized as an adaptive kernel regression:
\begin{equation}
H_i = \sum_j \kappa(\tilde{x}_i, \tilde{x}_j) \cdot x_j,
\end{equation}
where the kernel function \( \kappa(\cdot, \cdot) \) is implicitly defined by feature similarity. 
Since the features contain pose information, this kernel function is able to learn cross-view association patterns related to 3D geometric relationships during training, thereby promoting geometric consistency in the latent space.
Furthermore, if we define the cross-view feature consistency error as
\begin{equation}
\mathcal{L}_{cons} = \sum_{i,j} A_{ij} \|x_i - x_j\|^2,
\end{equation}
the above attention update is equivalent to a kernel smoothing operation, which can reduce this consistency error during optimization, thus achieving cross-view feature alignment.

At the same time, our method achieves the complementary integration of local and global information through the collaborative modeling of the geometric and semantic branches. 
Specifically, the local geometric branch tends to capture high-frequency spatial structures (such as depth discontinuities and boundaries), while the attention-based semantic branch models long-range dependencies by globally weighted aggregation, focusing more on low-frequency structures.
In summary, this design can be understood as learning a pose-conditioned kernel regression operator, which preserves local details while effectively enhancing cross-view consistency through global information interaction.

\section{Theoretical Analysis of the Dual-Branch Balanced CFG}
In conditional diffusion models, the introduction of a conditional variable \( c = (c_g, c_s) \) aims to learn the conditional distribution \( p(x \mid c) \), ensuring that the generated results maintain a statistical dependency with the condition variable. 
Although the training objective of the diffusion model does not directly maximize mutual information, the maximum likelihood objective can be interpreted as implicitly encouraging the increase of mutual information between \( x \) and the condition variables.
According to the chain rule for mutual information, we have:
\begin{equation}
I(x; c_g, c_s) = I(x; c_g) + I(x; c_s \mid c_g),
\end{equation}

In standard joint training, if the geometric condition \(c_g\) provides strong structural constraints, the model may preferentially utilize \(c_g\) to minimize the loss during the early stages of optimization, thereby weakening its reliance on the semantic condition \(c_s\). 
This phenomenon can be interpreted as an information shortcut, where the model tends to rely on the conditioning variable that is easier to optimize.
To mitigate this issue, we sample the condition \(c\) from the following set during training:
\begin{equation}
c \sim \{\emptyset, c_g, c_s, (c_g,c_s)\},
\end{equation}
and the corresponding optimization objective is defined as:
\begin{equation}
\mathcal{L} = \mathbb{E}_{c} \mathbb{E}_{x,\epsilon,t}[\|\epsilon - \epsilon_\theta(x_t,c,t)\|^2].
\end{equation}
This objective is equivalent to simultaneously fitting multiple conditional distributions:
\begin{equation}
p(x),\quad p(x\mid c_g),\quad p(x\mid c_s),\quad p(x\mid c_g,c_s).
\end{equation}
From the information-theoretic perspective, this training strategy explicitly encourages the model to maintain robust predictive capabilities across different conditional subspaces, thereby preventing the model from relying solely on a single conditioning variable and mitigating the issue of imbalanced information utilization.

\section{Discussion on Pre-trained Vision Models and Metric Scale}
\noindent\textbf{Replaceability of the Pretrained Vision Model.} 

From the perspective of the framework interface, LiFi does not depend on the specific internal architecture of DepthAnything3.
The geometric branch only requires the depth foundation model to provide two types of per-pixel information for each input view: depth predictions and their corresponding reliability estimates. 
LiFi's reverse sampling module then aligns these predictions with the target LiDAR scanning grid, while the uncertainty-aware encoder uses the reliability signals to modulate the geometric features. 
Therefore, as long as an alternative model can provide outputs in the same form, the subsequent inverse sampling, FiLM modulation, and diffusion generation pipeline can remain unchanged.

Therefore, from an architectural perspective, any large monocular depth model capable of estimating per-pixel uncertainty can technically be directly integrated into LiFi. 
However, if the model processes each camera view independently, the predicted depths may have independent scales across different views. 
In this case, we recommend introducing cross-view scale alignment before reverse sampling.

Specifically, one feasible approach is to use the known camera calibration and overlapping fields of view to perform scale-and-shift alignment across the depth maps from different views. 
For example, the prediction for each view can be expressed as:
$\hat{D_n} = a_n D_n + b_n$, where $a_n$ and $b_n$ are estimated by enforcing consistency within the overlapping regions. After cross-view alignment, a unified normalization scheme can be applied to all views before feeding them into LiFi.

Therefore, multi-view consistency is not a strict architectural requirement for LiFi, but it can positively contribute to maintaining the quality of the geometric conditions. 
In addition, methods with multi-view depth estimation capabilities, such as VGGT and MapAnything, can serve as direct replacements for DepthAnything3.
Therefore, DepthAnything3 is an instantiation choice that matches the task requirements, rather than an irreplaceable component of the LiFi framework.
\\

\noindent\textbf{Metric Scale of the Simulator.} 

We first emphasize that the final LiDAR generated by LiFi is already represented in the metric LiDAR coordinate system of the target dataset, rather than having only relative scale.

LiFi trains the diffusion model with range images constructed from real LiDAR scans as the target, and therefore learns a mapping from relative visual geometry and image semantics to real metric LiDAR distance distributions. 
After de-normalization, each valid cell in the generated range image represents an actual distance in the target sensor coordinate system. 
It can then be back-projected using the corresponding laser-beam elevation angles, azimuth sampling, and sensor calibration parameters to obtain a 3D point cloud in meters. 
The range image transformation adopted in the paper is itself reversible, and the generation target is constructed from the metric distances of real LiDAR scans.

Therefore, for datasets such as nuScenes and KITTI-360, there is no need to fit an additional global scale factor after generation. 
Metric LiDAR point clouds from datasets such as KITTI and Argoverse can likewise be directly converted into range images according to the scanning pattern of the target sensor and used as training supervision for LiFi.

If LiFi is to be adapted to a specific simulator, we recommend the following procedure:

1. If the simulator is intended to use a target LiDAR configuration consistent with those of current mainstream datasets, the LiFi architecture trained on these datasets can be directly used.

2. If the goal is to simulate a completely new virtual target LiDAR configuration, the first step is to specify the target LiDAR settings, including the beam elevation angles, horizontal angular resolution, and extrinsic parameters. 
Based on these settings, the target range image representation can then be constructed, and the VAE and diffusion model can be trained or fine-tuned using camera–metric LiDAR data corresponding to that sensor. 
At inference time, the generated results can be converted back into a metric point cloud according to the same normalization parameters and sensor model.

\section{Analysis of Computational Efficiency}
In this section, we provide an analysis of the inference efficiency of LiFi. 
As shown in Table \ref{table_time}, we present the performance and inference speed of our method across different denoising timesteps.

\begin{table}[ht]
\centering
\vspace{2mm}
\caption{Quantitative evaluation of LiFi's inference efficiency. The MMD values are reported in $10^{-5}$ and JSD in $10^{-2}$. Inference time is evaluated in s/frame.}
\vspace{3mm}
\label{table_time}
\setlength{\tabcolsep}{6pt}
\renewcommand{\arraystretch}{1.1}
\resizebox{0.5\columnwidth}{!}{
\small
\begin{tabular}{l|cccc} 
\toprule
\textbf{Denoising Step}  & \textbf{MMD} & \textbf{JSD} & \textbf{CD}  & \textbf{Time}\\
\midrule \midrule 
$ T_{\text{inference}} = 10$  & 180.97 & 18.00 & 1.29 & 0.11  \\
$ T_{\text{inference}} = 30$  & 0.98 & 4.29 & 1.00 & 0.31  \\
\rowcolor{myyellow}%
$ T_{\text{inference}} = 50$  & 2.78 & 3.66 & 0.83 & 0.52 \\
$ T_{\text{inference}} = 100$  & 2.73 & 3.84 & 0.83 & 1.04  \\
$ T_{\text{inference}} = 200$  & 2.76 & 3.83 & 0.82 & 2.08  \\
\bottomrule
\end{tabular}
}
\end{table}

\section{Ablation Study on the Number of Input Image Views}
The number of input RGB images does not affect the fairness of the experimental comparisons but is merely an architectural design choice. 
In this section, we present an ablation study on the nuScenes dataset using 1, 2, 4, and 6 input camera images, as shown in Table~\ref{table_num_images}.

\begin{table}[ht]
\centering
\vspace{2mm}
\caption{Quantitative evaluation of the impact of the number of input images on LiFi's generation performance. The MMD values are reported in $10^{-5}$ and JSD in $10^{-2}$.}
\vspace{3mm}
\label{table_num_images}
\setlength{\tabcolsep}{6pt}
\renewcommand{\arraystretch}{1.1}
\resizebox{0.45\columnwidth}{!}{
\small
\begin{tabular}{l|cccc}
\toprule
\textbf{Num.} & \textbf{MMD} $\downarrow$ & \textbf{JSD} $\downarrow$ & \textbf{CD} $\downarrow$ & \textbf{EMD} $\downarrow$ \\
\midrule \midrule
1 & 14.98 & 4.83 & 1.93 & 112.54 \\
2 & 7.46 & 4.35 & 1.53 & 99.48 \\
4 & 5.85 & 4.17 & 1.19 & 76.34 \\
\rowcolor{myyellow}%
6 & 2.78 & 3.66 & 0.83 & 50.08 \\
\bottomrule
\end{tabular}
}
\end{table}

\section{Ablation Study on the Range Image Resolution}
In this section, we conduct an ablation study on the resolution of the range images, with the results presented in Table~\ref{table_resolution}.
At a resolution of \([1024, 32]\), most pixels correspond to LiDAR points in the scene, whereas \([512, 32]\) results in the omission of approximately half of the points in the complete LiDAR scene during representation conversion. 
In contrast, \([2048, 32]\) introduces nearly an equal number of additional empty pixels, which can instead interfere with learning in valid regions.

\begin{table}[ht]
\centering
\caption{Quantitative evaluation of LiFi's generation performance and inference efficiency across different range image resolutions. The MMD values are reported in $10^{-5}$ and JSD in $10^{-2}$.}
\vspace{3mm}
\label{table_resolution}
\setlength{\tabcolsep}{6pt}
\renewcommand{\arraystretch}{1.3}
\resizebox{0.48\columnwidth}{!}{
\small
\begin{tabular}{l|ccc}
\toprule
\textbf{Range Image Size} & \textbf{MMD} $\downarrow$ & \textbf{JSD} $\downarrow$ & \textbf{Time} $\downarrow$ \\
\midrule \midrule
$[512, 32]$ & 740.77 & 31.04 & 0.29 \\
\rowcolor{myyellow}%
$[1024, 32]$ & 2.78 & 3.66 & 0.52 \\
$[2048, 32]$ & 3.32 & 4.03 & 0.97 \\
\bottomrule
\end{tabular}
}
\end{table}

\section{Quantitative Analysis of Failure Cases at Long Range}
In this section, we evaluate the generation performance of our model across different depth ranges.
As shown in Table~\ref{table_distance}, LiFi's performance gradually degrades as the distance increases.

\begin{table}[H]
\centering
\vspace{2mm}
\caption{Quantitative evaluation of LiFi's generation performance across different radial distance ranges. The MMD values are reported in $10^{-5}$ and JSD in $10^{-2}$.}
\vspace{3mm}
\label{table_distance}
\setlength{\tabcolsep}{6pt}
\renewcommand{\arraystretch}{1.3}
\resizebox{0.7\columnwidth}{!}{
\small
\begin{tabular}{l|ccccc}
\toprule
\textbf{Radial Distance} & \textbf{Proportion of Points} & \textbf{MMD} $\downarrow$ & \textbf{JSD} $\downarrow$ & \textbf{CD} $\downarrow$ & \textbf{EMD} $\downarrow$ \\
\midrule \midrule
$[0, 10)$ m      & 60.87\% & 2.08 & 2.06 & 0.11 & 2.05 \\
$[10, 20)$ m     & 23.69\% & 1.40 & 3.80 & 0.94 & 34.62 \\
$[20, 30)$ m     & 9.25\%  & 2.48 & 5.52 & 4.13 & 189.64 \\
$[30, 40)$ m     & 3.92\%  & 5.18 & 6.61 & 12.06 & 538.58 \\
$[40, +\infty)$ m & 2.27\% & 7.03 & 9.23 & 32.82 & 1264.71 \\
\bottomrule
\end{tabular}
}
\end{table}


\clearpage
\section{More Qualitative Results}
\begin{figure}[H]
  \centering
  \includegraphics[width=\linewidth]{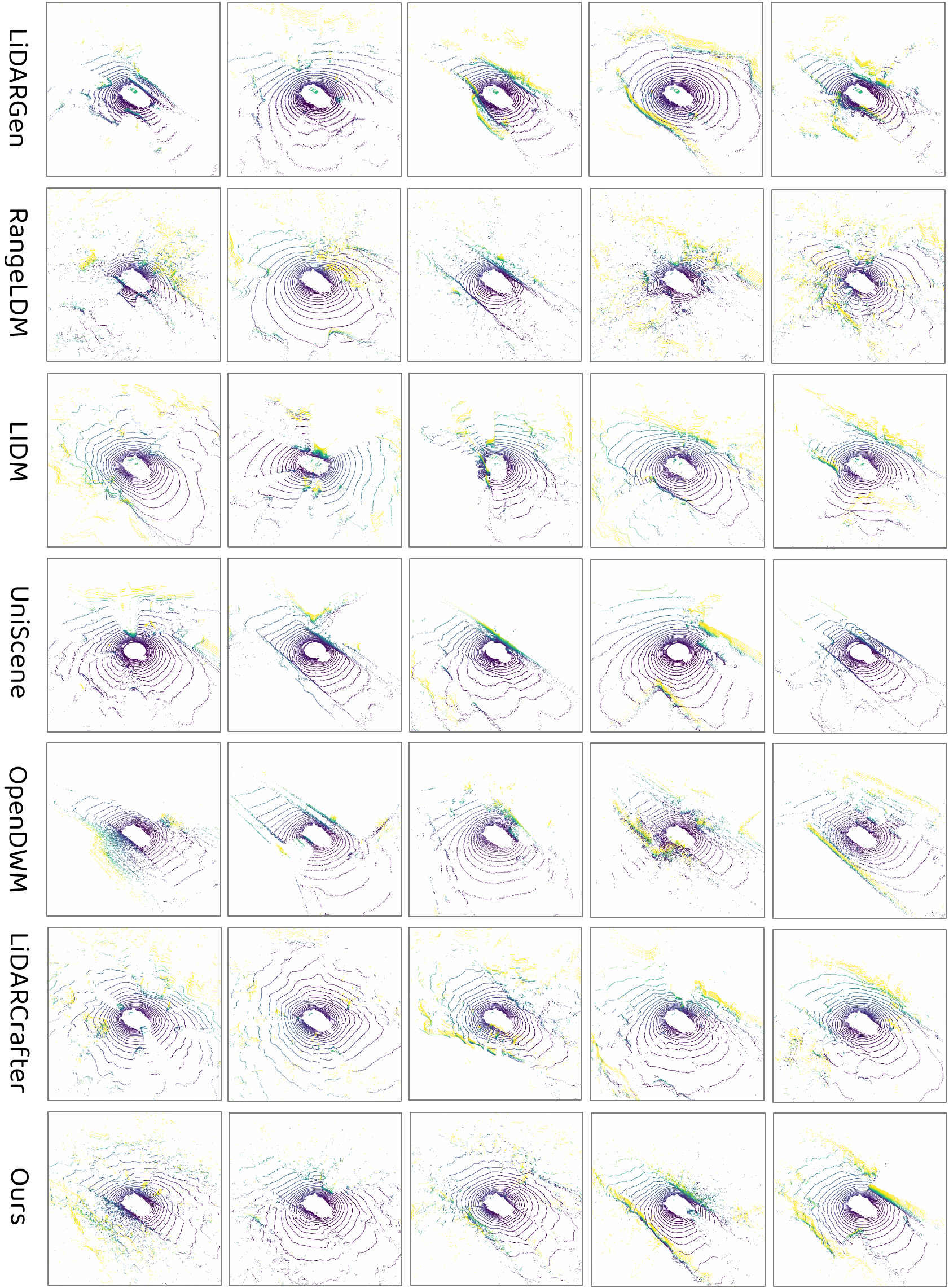}\\
  \caption{\label{more_gen_results} 
    Additional qualitative generation examples of our method and several baseline approaches on the LiDAR generation task on the nuScenes dataset.
}
\end{figure}

\begin{figure}[H]
  \centering
  \includegraphics[width=0.99\linewidth]{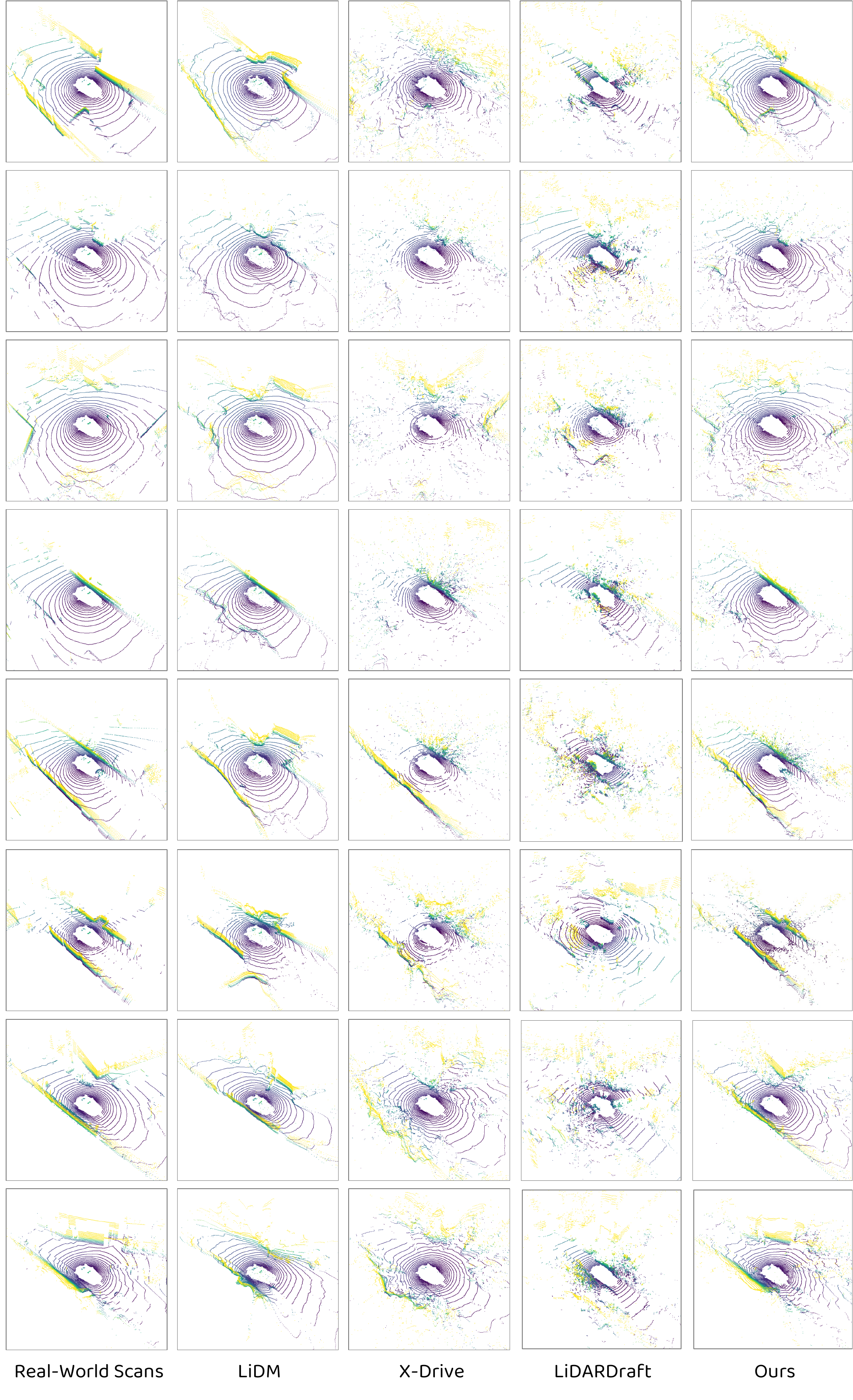}\\
  \vspace{-3mm}
  \caption{\label{more_results} 
    Qualitative comparisons of our method with several baseline approaches on the camera-to-LiDAR task across more representative scenarios on the nuScenes dataset.
}
\end{figure}

In this section, we first present additional visual examples comparing the generation results of our method with those of LiDAR generation baselines, as shown in Figure~\ref{more_gen_results}. 
We then provide more camera-to-LiDAR visualization results across diverse scenarios, as shown in Figure~\ref{more_results}.

\end{document}